\documentclass[10pt,twocolumn,letterpaper]{article}

\usepackage{cvpr}              

\usepackage[dvipsnames]{xcolor}

\definecolor{cvprblue}{rgb}{0.21,0.49,0.74}
\usepackage[pagebackref,breaklinks,colorlinks,citecolor=cvprblue]{hyperref}

\usepackage{booktabs}
\usepackage{multirow}
\usepackage{xcolor}
\usepackage{algorithm}
\usepackage{algorithmic}
\usepackage{multirow}
\usepackage[normalem]{ulem}

\usepackage{amsthm} 

\newtheorem{hypothesis}{Hypothesis}

\newtheorem{lemma}{Lemma}

\newtheorem{proposition}{Proposition}

\usepackage{graphicx}
\usepackage{subcaption}

\usepackage{multicol}
\usepackage{color,colortbl}

\definecolor{Gray}{gray}{0.45}
\newcolumntype{a}{>{\columncolor{Gray}}c}

\definecolor{my_blue}{rgb}{0.8431, 0.9300, 0.8431}
\definecolor{my_b}{rgb}{0.906, 0.945, 0.976}
\definecolor{my_g}{rgb}{0.898, 1.000, 0.898}
\definecolor{my_y}{rgb}{1.000, 0.973, 0.898}
\definecolor{my_gray}{rgb}{0.925, 0.925, 0.925}

\newcolumntype{b}{>{\columncolor{my_b}}c}
\newcolumntype{g}{>{\columncolor{my_g}}c}
\newcolumntype{y}{>{\columncolor{my_y}}c}
\newcolumntype{a}{>{\columncolor{my_gray}}c}
\definecolor{my_r}{rgb}{0.949, 0.474, 0.439}

\newcommand{\repeatthanks}{\textsuperscript{\thefootnote}}

\def\paperID{9788} 
\def\confName{CVPR}
\def\confYear{2024}

\title{MAP: MAsk-Pruning for Source-Free Model Intellectual Property Protection}

\author{Boyang Peng$^{1}$\thanks{Equal Contribution}, Sanqing Qu$^{1}$\repeatthanks, Yong Wu$^{1}$, Tianpei Zou$^{1}$, Lianghua He$^{1}$,\\
Alois Knoll$^{2}$, Guang Chen$^{1}$\thanks{Corresponding author: guangchen@tongji.edu.cn}\ , Changjun Jiang$^{1}$\\
{\small $^{1}$Tongji University,}
{\small $^{2}$ Technical University of Munich}\\
}

\begin{document}
\maketitle
\begin{abstract}

Deep learning has achieved remarkable progress in various applications, heightening the importance of safeguarding the intellectual property (IP) of well-trained models. It entails not only authorizing usage but also ensuring the deployment of models in authorized data domains, i.e., making models exclusive to certain target domains. Previous methods necessitate concurrent access to source training data and target unauthorized data when performing IP protection, making them risky and inefficient for decentralized private data. In this paper, we target a practical setting where only a well-trained source model is available and investigate how we can realize IP protection. To achieve this, we propose a novel MAsk Pruning (MAP) framework. MAP stems from an intuitive hypothesis, i.e., there are target-related parameters in a well-trained model, locating and pruning them is the key to IP protection. Technically, MAP freezes the source model and learns a target-specific binary mask to prevent unauthorized data usage while minimizing performance degradation on authorized data. Moreover, we introduce a new metric aimed at achieving a better balance between source and target performance degradation. To verify the effectiveness and versatility, we have evaluated MAP in a variety of scenarios, including vanilla source-available, practical source-free, and challenging data-free. Extensive experiments indicate that MAP yields new state-of-the-art performance. Code will be available at \url{https://github.com/ispc-lab/MAP}.

\end{abstract}
    
\section{Introduction}
\label{sec:intro}
\vspace{-0.20in}

\begin{figure}[t]
  \centering
   \includegraphics[width=\columnwidth]{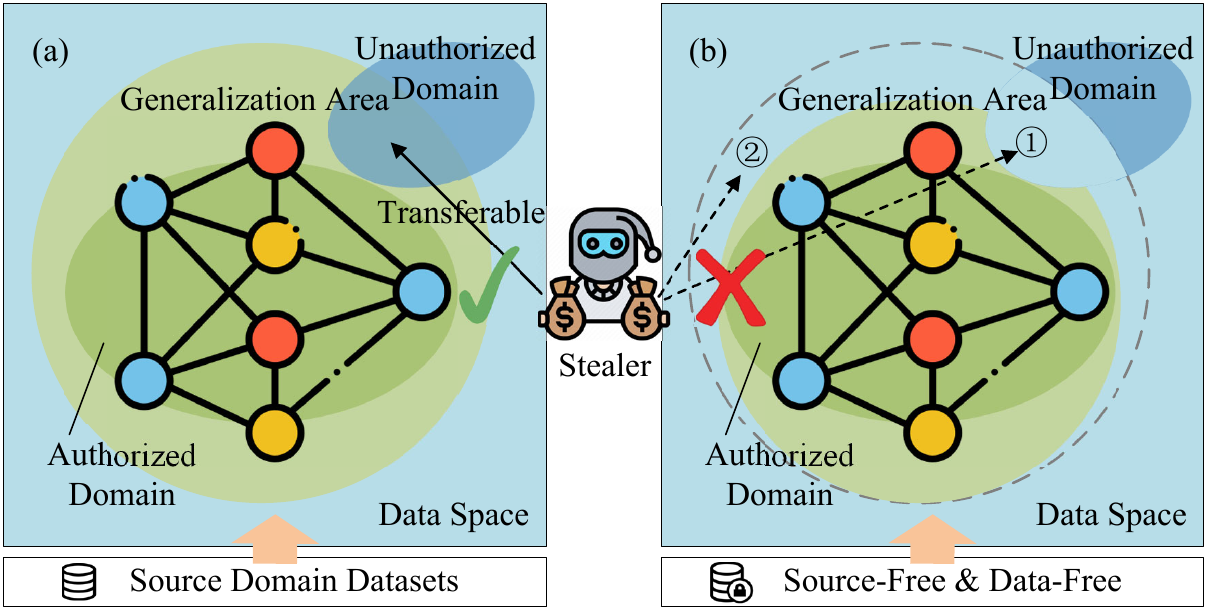}
   \vspace{-0.20in}
   \caption{An illustration of model IP protection in source-free and data-free situations. (\textbf{a}) The original model is well-trained in the authorized (source) domain, with a wide generalization area that allows illegal access to the model through unauthorized (target) domains. (\textbf{b}) Two methods are shown: (1) Source-free IP protection, which removes an unauthorized domain from the generalization area without using source datasets; and (2) Data-Free IP protection, which cannot access any datasets but reduces the generalization area, preventing illegal knowledge transfer.}
   \vspace{-0.25in}
   \label{Motivation}
\end{figure}

\hspace{1em} \par With the growing popularity of deep learning in various applications (such as autonomous driving, medical robotics, virtual reality, etc), the commercial significance of this technology has soared. However, obtaining well-trained models is a resource-intensive process. It requires considerable time, labor, and substantial investment in terms of dedicated architecture design~\cite{he2016deep, dosovitskiy2020_vit}, large-scale high-quality data~\cite{deng2009imagenet, kirillov2023_sam}, and expensive computational resources~\cite{zoph2016neural}. Consequently, safeguarding the intellectual property (IP) of well-trained models has received significant concern.

\par Previous studies on IP protection mainly focus on ownership verification~\cite{li2023plmmark, charette2022cosine,vybornova2022method} and usage authorization~\cite{peng2022fingerprinting,guan2022you}, i.e., verifying who owns the model and authorizing who has permission to use it. Despite some effectiveness, these methods are vulnerable to fine-tuning or re-training. Moreover, authorized users retain the freedom to apply the model to any data without restrictions. Consequently, they effortlessly transfer high-performance models to similar tasks, leading to hidden infringement. Therefore, comprehensive IP protection requires a thorough consideration of applicability authorization. It entails not only authorizing usage but also preventing the usage of unauthorized data.

\par The primary challenge lies in the fact that the generalization region of well-trained models typically encompasses some unauthorized domains (as depicted in Fig.~\ref{Motivation} (a)). It arises from the models' innate ability to capture domain-invariant features, thereby leading to potential applicability IP infringements. An intuitive solution is to make the generalization bound of models more explicit and narrower, i.e., optimizing models to prioritize domain-dependent features and confining their applicability exclusively to authorized domains. To achieve this, NTL~\cite{wang2021non} first remolds the methodology in domain adaptation with an opposite objective. It amplifies the maximum mean difference (MMD) between the source (authorized) and target (unauthorized) domains, thereby effectively constraining the generalization scope of the models. Different from NTL, CUTI-domain~\cite{wang2023model} constructs middle domains with combined source and target features, which then act as barriers to block illegal transfers from authorized to unauthorized domains. Regardless of promising results, these methods require concurrent access to both source and target data when performing IP protection, rendering them unsuitable for decentralized private data scenarios. Moreover, they typically demand retraining from scratch to restrict the generalization boundary, which is highly inefficient since we may not have prior knowledge of all unauthorized data at the outset, leading to substantial resource waste.

\par In this paper, we target a practical but challenging setting where a well-trained source model is provided instead of source raw data, and investigate how we can realize source-free IP protection. To achieve this, we first introduce our \textit{Inverse Transfer Parameter Hypothesis} inspired by the {lottery ticket hypothesis}~\cite{frankle2018lottery}. We argue that well-trained models contain parameters exclusively associated with specific domains. Through deliberate pruning of these parameters, we effectively eliminate the generalization capability to these domains while minimizing the impact on other domains. To materialize this idea, we propose a novel MAsk Pruning (\textbf{MAP}) framework. MAP freezes the source model and learns a target-specific binary mask to prevent unauthorized data usage while minimizing performance degradation on authorized data. For a fair comparison, we first compare our MAP framework with existing methods when source data is available, denoted as SA-MAP. Subsequently, we evaluate MAP in source-free situations. Inspired by data-free knowledge distillation, we synthesize pseudo-source samples and amalgamate them with target data to train a target-specific mask for safeguarding the source model. This solution is denoted as SF-MAP. Moreover, we take a step further and explore a more challenging data-free setting, where both source and target data are unavailable. Building upon SF-MAP, we introduce a diversity generator for synthesizing auxiliary domains with diverse style features to mimic unavailable target data. We refer to this solution as DF-MAP. For performance evaluation, current methods only focus on performance drop on target (unauthorized) domain, but ignore the preservation of source domain performance. To address this, we introduce a new metric Source \& Target Drop (\textit{ST-D}) to fill this gap. We have conducted extensive experiments on several datasets, the results demonstrate the effectiveness of our MAP framework. The key contributions are summarized as follows:

\begin{itemize}
        \item To the best of our knowledge, we are the first to exploit and achieve source-free and data-free model IP protection settings. These settings consolidate the prevailing requirements for both model IP and data privacy protection.

        \item We propose a novel and versatile MAsking Pruning (\textbf{MAP}) framework for model IP protection. MAP stems from our \textit{Inverse Transfer Parameter Hypothesis}, i.e., well-trained models contain parameters exclusively associated with specific domains, pruning these parameters would assist us in model IP protection.

        \item Extensive experiments on several datasets, ranging from source-available, source-free, to data-free settings, have verified and demonstrated the effectiveness of our MAP framework. Moreover, we introduce a new metric for thorough performance evaluation.

\end{itemize}

\section{Related Work}
\label{sec:formatting}

\begin{figure*}[t]
  \centering
   \vspace{-0.15in}
   \includegraphics[width=17.5cm]{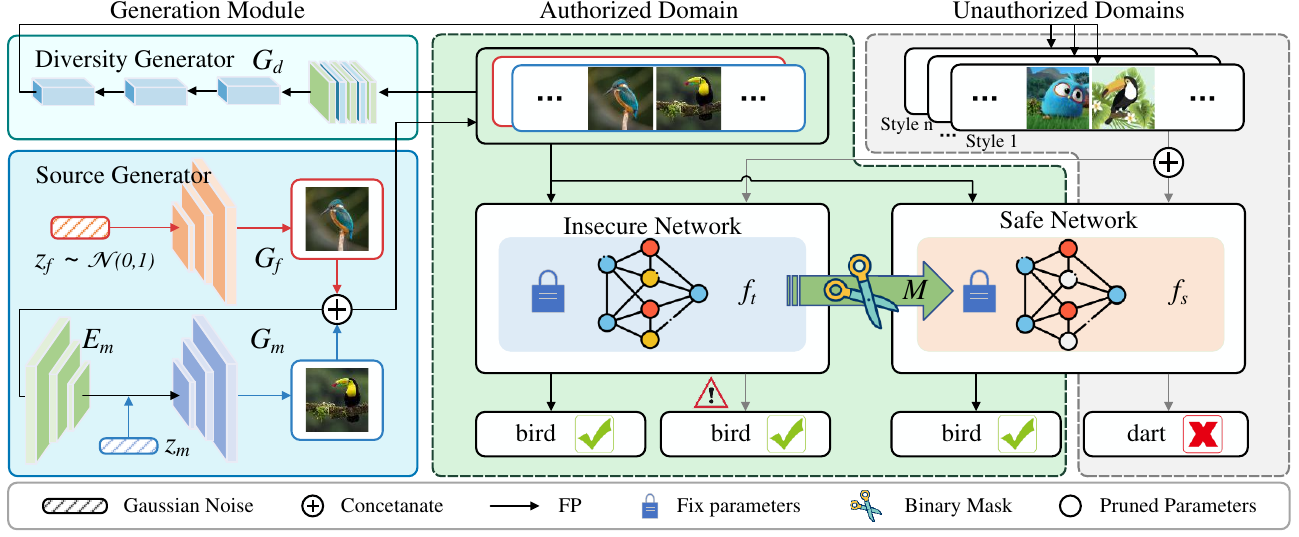}
   \vspace{-0.25in}
   \caption{Overall architecture of MAP. Please note that this architecture presents the complete DF-MAP, from which SA-MAP and SF-MAP are derived. \textbf{(a)} The Generation Module, displayed in the left part, consists of three generators. The \textit{Diversity Generator ($G_d$)} synthesizes auxiliary samples to generate neighbor domains with multiple style features. The \textit{Fresh Generator ($G_f$)} generates synthetic novel featured samples, while \textit{Memory Generator ($G_m$)} replays samples with features from previous images. In SF-MAP, the \textit{Diversity Generator ($G_d$)} is removed, and existing target domain data is utilized for training. In SA-MAP, the entire Generation Module is eliminated, and existing source domain data is further leveraged, as detailed in the supplementary material. \textbf{(b)} The right part illustrates the mask-pruning process. A well-trained original source network $f_s$ distills knowledge into the target network $f_t$, which shares the same architecture. We initialize and fix them with the same checkpoint, then update a \textit{Learnable Binary Mask ($M$)} with consistency loss calculated from synthetic samples. The MAP limits a target domain generalization region while retaining source domain performance, leading to a beneficial outcome.}
   \vspace{-0.20in}
   \label{fig:df-architecture}
\end{figure*}

\par \hspace{1em} \textbf{Model Intellectual Property Protection.} To gain improper benefits and collect private information in the model, some individuals have developed attack methods, such as the inference attack~\cite{carlini2022membership, liu2022membership, ye2022enhanced}, model inversion attack~\cite{fredrikson2015model,zhang2022model,struppek2022plug}, adversarial example attack~\cite{luo2018towards,yuan2021current} and others~\cite{binici2022robust, lopes2017data}. Therefore, the protection of model intellectual property rights has become important. Recent research has focused on ownership verification and usage authorization to preserve model intellectual property~\cite{wang2021non}. Traditional ownership verification methods \cite{zhang2018protecting,li2019prove} employ watermarks to establish ownership by comparing results with and without watermarks. However, these techniques are also susceptible to certain watermark removal~\cite{guo2020fine, chen2021refit} techniques. Usage authorization typically involves encrypting the entire or a portion of the network using a pre-set private key for access control purposes~\cite{peng2022fingerprinting,guan2022you}.

\par As a usage authorization method, NTL~\cite{wang2021non} builds an estimator with the characteristic kernel from Reproduction Kernel Hilbert Spaces (RKHSs) to approximate the Maximum Mean Discrepancy (MMD) between two distributions to achieve the effect of reducing generalization to a certain domain. According to~\cite{wang2023model}, CUTI-Domain generates a middle domain with source style and target semantic attributes to limit generalization region on both the middle and target domains. \cite{wang2023domain} enhances network performance by defining a divergence ball around the training distribution, covering neighboring distributions, and maximizing model risk on all domains except the training domain. However, the privacy protection policy results in difficulties in getting user source domain database data, which disables the above methods. To address this challenge, in this paper, we propose source-free and target-free model IP protection tasks.

\par \textbf{Unstructured Parameter Pruning.} Neural network pruning reduces redundant parameters in the model to alleviate storage pressure. It is done in two ways: unstructured or structured~\cite{he2023structured}. Structured pruning removes filters, while unstructured pruning removes partial weights, resulting in fine-grained sparsity. Unstructured pruning is more effective, but sparse tensor computations save runtime, and compressed sparse row forms add overhead~\cite{wimmer2022interspace}.

\par There are several advanced methods for unstructure pruning. For example, \cite{mallya2018piggyback} proposes a method that builds upon the concepts of network quantization and pruning, which enhances the network performance for a new task by utilizing binary masks that are either applied to unmodified weights on an existing network. As the basis of our hypothesis, the lottery hypothesis~\cite{frankle2018lottery} illustrates that a randomly initialized, dense neural network has a sub-network that matches the test accuracy of the original network after at most the same number of iterations trained independently. 
\par \textbf{Source-Free Domain Adaptation.} Domain adaptation addresses domain shifts by learning domain-invariant features between source and target domains~\cite{wang2018deep}. In terms of source-free learning, our work is similar to source-free domain adaptation, which is getting more attention due to the data privacy policy~\cite{liang2023comprehensive}. \cite{chidlovskii2016domain} first considers prevention access to the source data in domain adaptation and then adjusts the source pre-trained classifier on all test data. Several approaches are used to apply the source classifier to unlabeled target domains~\cite{clinchant2016transductive, van2017unsupervised}, and the current source-free adaptation paradigm does not exploit target labels by default~\cite{liang2020we, liu2021ttt++, sanqing2024_LEAD}. Some schemes adopt the paradigm based on data generation~\cite{li2020model, qiu2021source, liu2021source}, while others adopt the paradigm based on feature clustering~\cite{liang2021source, liang2020we, qu2022bmd, sanqing2023_GLC}. Our technique follows the former paradigm, which synthesizes a pseudo-source domain with prior information. 

\section{Methodology}

\par \hspace{1em} In this paper, we designate the source domain as the authorized domain and the target domain as the unauthorized domain. Firstly, Section~\ref{method:hypothesis} defines the problem and our \textit{Inverse Transfer Parameter Hypothesis}. Then Section~\ref{method:sa-ntl}, Section~\ref{method:sf-ntl}, and Section~\ref{method:df-ntl} present our MAsk-Pruning (MAP) framework in source-available, source-free, and data-free situations, respectively. 

\subsection{Problem Definition}
\label{method:hypothesis}
\par \hspace{1em} Formally, we consider a source network $f_s:\mathcal{X}_s\rightarrow \mathcal{Y}_s$ trained on the source domain $\mathcal{D}_s=\{(x_s,y_s)||x_s\sim \mathcal{P}_\mathcal{X}^s, y_s\sim \mathcal{P}_\mathcal{Y}^s\}$, a target network $f_t:\mathcal{X}_t\rightarrow \mathcal{Y}_t$, and target domain $\mathcal{D}_t=\{(x_t,y_t)||x_t\sim \mathcal{P}_\mathcal{X}^t, y_t\sim \mathcal{P}_\mathcal{Y}^t\}$. $\mathcal{P}_\mathcal{X}$ and $\mathcal{P}_\mathcal{Y}$ are the distribution of $\mathcal{X}$ and $\mathcal{Y}$, respectively. The goal of \textbf{\textit{source-available IP protection}} is to fine-tune $f_t$ while minimizing the generalization region of $f_t$ on target domain $\mathcal{D}_t$ by using $f_s$ with $\{x_s^i,y_s^i\}_{i=1}^{N_s}$ and $\{x_t^i,y_t^i\}_{i=1}^{N_t}$, in other words, degrade the performance of $f_t$ on $\mathcal{D}_t$ while preserving its performance on $\mathcal{D}_s$~\cite{wang2021non,wang2023model}. Due to increasingly stringent privacy protection policies, access to the $\mathcal{D}_s$ or $\mathcal{D}_t$ database of a user is more and more difficult~\cite{liang2023comprehensive}. Thus, we introduce IP protection for source-free and data-free scenarios. The objective of \textbf{\textit{source-free model IP protection}} is to minimize the $f_t$ generalization region of a designated target domain $\mathcal{D}_t$ by utilizing $f_s$ with $\{x_t^i\}_{i=1}^{N_t}$. \textbf{\textit{Data-free model IP protection}} is an extreme case. The objective is to minimize the generalization bound of $f_t$ by solely utilizing $f_s$, without access to $\mathcal{D}_s$ and $\mathcal{D}_t$. 

\par To mitigate the risk of losing valuable knowledge stored in the model parameters, we initiate our approach with unstructured pruning of the model. The lottery ticket hypothesis, proposed by \cite{frankle2018lottery}, is widely acknowledged as a fundamental concept in the field of model pruning. Building upon this foundation, we extend our \textit{Inverse Transfer Parameter Hypothesis} as Hypothesis~\ref{hypothesis:Inverse} in alignment with the principles presented in~\cite{frankle2018lottery}.

\begin{hypothesis}[\textit{Inverse Transfer Parameter Hypothesis}]
\label{hypothesis:Inverse}
For a dense neural network $f_s$ well-trained on the source domain $\mathcal{D}_s$, there exists a sub-network $f_{sub}$ like this: while $f_{sub}$ achieves the same test accuracy as $f_s$ on $\mathcal{D}_s$, its performance significantly degrades on the target domain $\mathcal{D}_t$. The pruned parameters of $f_s$ relative to $f_{sub}$ are crucial in determining its generalization capacity to $\mathcal{D}_t$.

\end{hypothesis}

\subsection{Source-Avaliable Model IP Protection}
\label{method:sa-ntl}
\par \hspace{1em} To verify the soundness of Hypothesis~\ref{hypothesis:Inverse}, we first design the source-available MAsk Pruning (SA-MAP). $f_t$ has the same architecture as $f_s$ and is initialized with a well-trained checkpoint of $\mathcal{D}_s$. To maximize the risk of the target domain and minimize it in the source domain, we prune the $f_t$'s parameters by updating a binary mask $M(\theta_M)$ of it and get a sub-network $f_{sub}$ by optimizing the objective:
\begin{equation}
\begin{aligned}
\mathcal{L}_{SA}(f_t;\mathcal{X}_s,\mathcal{Y}_s, \mathcal{X}_t, \mathcal{Y}_t)= \frac{1}{N_s} \sum_{i=1}^{N_s} KL\left(p_t^S \| y_s\right) \\- min\{\lambda \cdot\frac{1}{N_t} \sum_{i=1}^{N_t} KL\left(p_t^T \| y_t\right), \alpha\}
\end{aligned}
\label{eq:sl_loss}
\end{equation}
where $K L(\cdot)$ presents the Kullback-Leibler divergence, $\{x_s^i,y_s^i\}_{i=1}^{N_s}$ and $\{x_t^i,y_t^i\}_{i=1}^{N_t}$ mean $N_s/N_t$ data and labels sampled from source domain $\mathcal{D}_s$ and target domain $\mathcal{D}_t$, respectively. $p_t^S = f_t(x_s)$, and $p_t^T = f_t(x_t)$. $\alpha$ and $\lambda$ are the upper bound and scaling factor, respectively, which aim to limit the over-degradation of domain-invariant knowledge. We set $\alpha=1.0$ and $\lambda=0.1$. 

\begin{algorithm}[t]
\caption{SF-MAP in Source-Free Model IP Protection}
\label{alg:sfmntl}
\begin{algorithmic}[1]
\REQUIRE The target dataset $\mathcal{X}_t$, source network $f_s(x;\theta_s)$, target network $f_t(x;\theta_t)$, pre-trained model parameters $\theta_0$, fresh generator $G_f(z;\theta_f)$, memory generator $G_m(z;\theta_m)$, encoder $E_m(x;\theta_e)$, mask $M(\theta_M)$, gaussian noise $z_f$ and $z_m$.
\STATE Initialize $\theta_s$ and $\theta_t$ with $\theta_0$ and fix them 
\WHILE{not converged}
    \STATE Generate sample $x_f = G_f(z_f)$, $x_m = G_m(z_m)$ 
    \STATE Update $\theta_f$ by $x_f$ as Eq.~(\ref{eq:train_fresh})
    \STATE Concatenate synthetic data $x_s^{\prime}$ by $x_f$ and $x_m$
    \STATE Update $\theta_m$ and $\theta_e$ by $x_s^{\prime}$ as Eq.~(\ref{eq:train_memeory})
    \STATE Update $\theta_M$ using $x_s^{\prime}$, and $x_t$ as Eq.~(\ref{eq:sf_loss})
\ENDWHILE
\RETURN Learned mask parameters $\theta_M$
\end{algorithmic}
\end{algorithm}

\subsection{Source-Free Model IP Protection}
\label{method:sf-ntl}
\par \hspace{1em} Under the source-free setting, we have no access to $\{x_s^i,y_s^i\}_{i=1}^{N_s}$. To address this, we construct a replay-based source generator module to synthesize $N_s$ pseudo-source domain data $\{x_s^{i\prime},y_s^{i\prime}\}_{i=1}^{N_s}$. As illustrated in Fig.~\ref{fig:df-architecture}, SF-MAP is deformed by removing the Diversity Generator module and employing unlabeled target data. 
\par Source generator module is composed of two generators to synthesize source feature samples, the fresh generator $G_f$ synthesizes samples with novel features, and the memory generator $G_m$ replays samples with origin features. Before sampling, we first train the fresh generator $G_f$ as Eq.~(\ref{eq:train_fresh}). The objective of $G_f$ is to bring novel information to $f_t$. To make $G_f$ synthesize the source-style samples, we leverage the loss function of Eq.~(\ref{eq:train_fresh}). The first two items of Eq.~(\ref{eq:train_fresh}) derive from~\cite{chen2019data}, called predictive entropy and activation loss terms. These terms are designed to encourage the generator to produce high-valued activation maps and prediction vectors with low entropy, that is, to keep the generated samples consistent with the characteristics of the origin samples. As for the third item, $JS$ denotes the Jensen-Shannon divergence, encouraging $f_t$ to obtain consistent results with $f_s$. 

\vspace{-0.20in}
\begin{equation}
\begin{aligned}
\mathcal{L}_f = \frac{1}{N} \sum_{i=1}^N\left[\lambda_1 t_T^i \log \left(p_s^{i\prime}\right)-\lambda_2 \mathcal{H}\left(p_s^{\prime}\right) + J S\left(p_s^{\prime} \| p_t^{\prime}\right) \right]
\end{aligned}
\label{eq:train_fresh}
\end{equation}
where $p_s^{\prime} = f_s(x_f)$, and $p_t^{\prime} = f_t(x_f)$. $x_f = G_f(z_f)$ means the generated novel sample by a gaussian noise $z_f \sim \mathcal{N}(0,1)$, while $t_T^i = argmax\left(p_s^{i\prime}\right)$. $\mathcal{H}\left(\cdot\right)$ denotes the entropy of the class label distribution. 

\par Along with the fresh generator $G_f$, we optimize the memory generator $G_m$ and encoder $E_m$ as Eq.~(\ref{eq:train_memeory}), which aims to replay the features from earlier distributions. To preserve the original features, we utilize the L1 distance to measure the similarity between the generated samples and reconstructed samples. The loss is defined as follows:

\vspace{-0.20in}
\begin{equation}
\begin{aligned}
\mathcal{L}_m = \frac{1}{N} \sum_{i=1}^N [||x_s^{\prime}-x_{re}||_1 + \sum_{l \in L}\left||f_s(x_s^{\prime})_l-f_s(x_{re})_l|\right|_1]
\end{aligned}
\label{eq:train_memeory}
\end{equation}
where $||\cdot||_1$ means the L1 distance. $L$ is the selected layer set of $f_s$. $x_{re} = D_m(E_m(x_s^{\prime}))$ means reconstructed sample from the encoder-decoder structure. $x_s^{\prime}$ denotes the input synthetic sample concatenated by $x_f$ and $x_m$. $x_m = G_m(z_m)$ means memory sample, and $z_m \sim \mathcal{N}(0,1)$. 

\par After the generation process, we choose unlabeled target samples $\{x_t^i\}_i^{N_t}$ and synthetic source samples $\{x_s^{i\prime}\}_i^{N_s^{\prime}}$ to train the target model $f_t$ based on Hypothesis~\ref{hypothesis:Inverse}, which is detailed in Algorithm~\ref{alg:sfmntl}. Adhering to the procedure in SA-MAP, we still employ the binary mask pruning strategy and update a $M(\theta_M)$ of $f_t$ as follows: 

\vspace{-0.30in}
\begin{equation}
\begin{aligned}
\mathcal{L}_{SF}(f_t, f_s;\mathcal{X}_s^{\prime},\mathcal{X}_t)= \frac{1}{N_s^{\prime}} \sum_{i=1}^{N_s^{\prime}} KL\left(p_t^{\prime\prime} \|~p_s^{\prime\prime}\right) \\ - min\{\lambda \cdot\frac{1}{N_t} \sum_{i=1}^{N_t} K L\left(p_t^T \| y_{psd}\right), \beta\} \\
y_{psd}= \begin{cases}f_s(x_{t}), & \text { if } \text {conf($p_s^{T}$)} > \Delta \\ \frac{1}{n} \sum_i f_s(Aug_i\left(x_t\right)), & \text { otherwise,}\end{cases}
\end{aligned}
\label{eq:sf_loss}
\end{equation}
where $p_s^{\prime\prime} = f_s(x_s^{\prime})$, $p_t^{\prime\prime} = f_t(x_s^{\prime})$, $p_t^{T} = f_t(x_t)$, $p_s^{T} = f_s(x_t)$, respectively. To improve pseudo labels $y_{psd}$, we utilize a set of $i$ data augmentation $Aug_i$, when the prediction confidence conf($p_s^{T}$) is lower than a threshold $\Delta$. $\lambda$ and $\beta$ denote a scaling factor and an upper bound, respectively. We set $\lambda = 0.1$ and $\beta = 1.0$.

\subsection{Data-Free Model IP Protection}
\label{method:df-ntl}
\par \hspace{1em} When faced with challenging data-free situations, we adopt an exploratory approach to reduce the generalization region. Inspired by~\cite{wang2021learning}, we design a diversity generator $G_d$ with learnable mean shift $\theta_{\mu}$ and variance shift $\theta_{\sigma}$ to extend the pseudo source samples to neighborhood domains $D_{nbh}$ of variant directions. The objective is to create as many neighboring domains as feasible to cover the most target domains and limit the generalization region. After generation, we concatenate samples with distinct directions as the whole pseudo auxiliary domain. 
\par The latent vector $z_i$ of $i$-th sample $x_i$ from the dataset $\mathcal{X}$ is generated by the feature extractor $g_s:\mathcal{X}\rightarrow \mathbb{R}^d$ of the source model $f_s = h_s(g_s(x))$, where $h_s: \mathbb{R}^d\rightarrow \mathbb{R}^k$ means the classifier, $d$ and $k$ mean the dimension of latent space and class number, respectively. This component is designed to learn and capture potential features for the generator $G_d$ in a higher-dimensional feature space. Eq.~(\ref{eq:mutual_information}) applies mutual information (MI) minimization to ensure variation between produced and original samples, ensuring distinct style features. 

\vspace{-0.20in}
\begin{equation}
\begin{aligned}
\hspace{-0.1cm}
\vspace{0.5cm}
\mathcal{L}_{MI}=\frac{1}{N} \sum_{i=1}^N [\log q\left(z_i^{\prime} \mid z_i\right)-\frac{1}{N} \sum_{j=1}^N \log q\left(z_j^{\prime} \mid z_i\right)] 
\end{aligned}
\label{eq:mutual_information}
\end{equation}

However, the semantic information between the same classes should be consistent. So we enhance the semantic consistency by minimizing the class-conditional maximum mean discrepancy (MMD)~\cite{sriperumbudur2010hilbert} in the latent space to enhance the semantic relation of the origin input sample $x$ and the generated sample $x_g$ in Eq.~(\ref{eq:semantic_consistency}), and $x_g = G_d(x)$.  
\vspace{-0.05in}
\begin{equation}
\begin{aligned}
\mathcal{L}_{sem}=\frac{1}{C} \sum_{k=1}^C[\|\frac{1}{N_k} \sum_{i=1}^{N_k} \phi\left(z_i^k\right)-\frac{1}{N_k^{\prime}} \sum_{j=1}^{N_k^{\prime}} \phi\left(z_j^{k\prime}\right)\|^2]
\end{aligned}
\label{eq:semantic_consistency}
\end{equation}
where $z^k$ and $z^{k\prime}$ denote the latent vector of class $k$ of $x$ and $x_g$. $q(\cdot)$ means an approximate distribution and $\phi(\cdot)$ means a kernel function. $N_k$ and $N_k^{\prime}$ are the number of origin and generated samples for class $k$, respectively. 


\par To generate diverse feature samples, we constrain different generation directions $n_{dir}$ of the gradient for the generation process detailed in supplementary material. The optimization process follows the gradient because it is the most efficient way to reach the goal. In this case, all the generated domains will follow the same gradient direction~\cite{wang2021non}. So we restrict the gradient to get neighborhood domains with diverse directions. We split the generator network $G_d$ into $n_{dir}$ parts. We limit direction $i$ by freezing the first $i$ parameters of convolutional layers. The gradient of the convolutional layer parameters is frozen during training, limiting the model's learning capabilities in that direction.

\section{Experiments}
\begin{figure*}[t]
  \centering
  \vspace{-0.2in}
   \includegraphics[width=16cm]{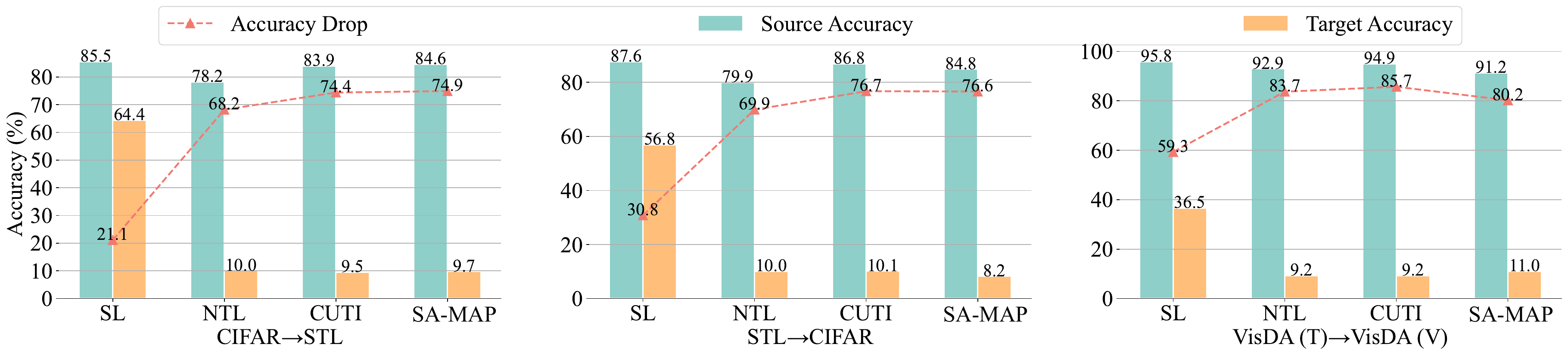}
   \vspace{-0.10in}
   \caption{The accuracy of SL, NTL, CUTI, and SA-MAP on CIFAR10$\rightarrow$STL10, and VisDA-2017 (T$\rightarrow$V). The `$\rightarrow$' represents the source domain transfer to the target domain. And the \textcolor[HTML]{8ECFC9}{green bar}, \textcolor[HTML]{FFBE7A}{orange bar} and \textcolor[HTML]{F27970}{red line} present the accuracy of the corresponding methods in the source domain, target domain, and relative degradation (Source Accuracy - Target Accuracy), respectively.} 
   \vspace{-0.15in}
   \label{fig:cifar_sl}
\end{figure*}

\subsection{Implementation Details}
\par \hspace{1em} \textbf{Experiment Setup.} Building upon existing works, we select representative benchmarks in transfer learning\textemdash the digit benchmarks (MNIST (MN)~\cite{deng2012mnist}, USPS (US)~\cite{hull1994database}, SVHN (SN)~\cite{netzer2011reading}, MNIST-M (MM)~\cite{ganin2016domain}) and CIFAR10~\cite{krizhevsky2009learning}, STL10~\cite{coates2011analysis} VisDA-2017~\cite{peng2017visda} for object recognition. For IP protection task, we employ the VGG11~\cite{simonyan2014very}, VGG13~\cite{simonyan2014very}, and VGG19~\cite{simonyan2014very} backbones, which is the same as~\cite{wang2023model}. The ablation study additionally evaluates on ResNet50~\cite{he2016deep}, ResNet101~\cite{he2016deep}, SwinT~\cite{liu2021swin} and Xception~\cite{chollet2017xception} backbones. We mainly compare our MAP with the NTL~\cite{wang2021non} and CUTI~\cite{wang2023model} baselines. We leverage the unitive checkpoints trained on supervised learning (SL) to initialize. To fairly compare in the source-free scenario, we replace the source and target data with synthetic samples with the generator in Section~\ref{method:sf-ntl} for all baselines. Experiments are performed on Python 3.8.16, PyTorch 1.7.1, CUDA 11.0, and NVIDIA GeForce RTX 3090 GPU. For each set of trials, we set the learning rate to 1e-4, and the batch size to 32. 
\par \textbf{Evaluation Metric.} Existing works~\cite{wang2021non,wang2023model} leverage the \textit{Source/Target Drop} metric ($Drop_{s}/Drop_{t}$), by quantifying the accuracy drop in the processed model compared to the original source model $f_s$ accuracy ($Acc_{s}/Acc_{t}$), to verify the effectiveness. However, these two separate metrics make it difficult to evaluate the effectiveness of the method as a whole because performance degradation on illegal target domains has the risk of destroying source domain knowledge. In order to realize the trade-off, we propose \textit{ST-D} as Eq.~(\ref{eq:st-d}). A lower \textit{ST-D} denotes enhanced IP protection, with a minimal drop on the source domain and a maximum on the target.
\begin{equation}
\begin{gathered}
\text{\textit{ST-D}}= \frac{Drop_{s}~/~Acc_{s}}{Drop_{t}~/~Acc_{t}}
\end{gathered}
\label{eq:st-d}
\end{equation}

\subsection{Result of MAP in Source-Available Situation}
\label{experiment:SA-MAP}
\par \hspace{1em} We first conduct experiments in the source-available situation to verify the effectiveness of our Hypothesis~\ref{hypothesis:Inverse}. As stated before, 

we introduce the SA-MAP and optimize a binary mask to realize model IP protection and compare NTL, CUTI, and SA-MAP on digit datasets. Results in Table~\ref{tab:supervised} show that SA-MAP achieves better Source Drop (-0.3\%) and \textit{ST-D} (-0.004), indicating the most deterioration in the target domain and the least in the source domain. It is noteworthy that SA-MAP even outperforms the origin model in the source domain (-0.3\%). We speculate this may be the result of pruning, which removes redundant parameters.
\par Fig.~\ref{fig:cifar_sl} assess method performance on CIFAR10, STL10, and VisDA-2017 benchmarks independently. We exploit VGG13 on CIFAR10$\rightarrow$STL10 experiment, and VGG19 on VisDA-2017 (T$\rightarrow$V). For a fair comparison, we utilize the same training recipe in \cite{wang2023model}.

A clear observation is that the pre-trained source model has good generalization performance on these three unauthorized tasks, seriously challenging the model IP. After performing protection, both baseline methods and our SA-MAP effectively reduce the performance on the target domain. SA-MAP achieves comparable or better results compared to baseline methods, which basically demonstrates our Hypothesis~\ref{hypothesis:Inverse}. 

\begin{table}[]
\addtolength{\tabcolsep}{0.0pt}
\resizebox{\columnwidth}{!}{%
\begin{tabular}{@{}c|c|cca}
\toprule
\multicolumn{1}{l|}{Methods} & Soure           & Source Drop$\downarrow$ & \multicolumn{1}{l}{Target Drop$\uparrow$} & \multicolumn{1}{l}{\textit{ST-D}$\downarrow$} \\ \midrule
\multirow{5}{*}{~NTL~\cite{wang2021non}}       & MT             & 1.5 (1.5\%)             & 50.9 (77.6\%)                             & 0.019                                 \\
                           & US              & \textbf{-0.2 (-0.2\%)}  & \textbf{46.3 (84.0\%)}                    & \textbf{-0.024}                       \\
                           & SN             & 0.8 (0.9\%)             & \textbf{50.0 (85.2\%)}                    & 0.011                                 \\
                           & MM             & 2.0 (2.1\%)             & 59.7 (79.2\%)                             & 0.027                                 \\ \cmidrule(l){2-5} 
                           & Mean            & 1.0  (1.1\%)            & \textbf{51.7 (81.5\%)}                    & 0.013                                 \\ \midrule
\multirow{5}{*}{~CUTI~\cite{wang2023model}}      & MT             & 0 (0\%)                 & \textbf{52.7 (80.0\%)}                    & 0                                     \\
                           & US             & -0.1 (-0.1\%)           & 42.3 (78.6\%)                             & -0.013                                \\
                           & SN              & 0.3 (0.3\%)             & 48.3 (82.3\%)                             & 0.036                                 \\
                           & MM               & 0.8 (0.8\%)             & 60.1 (80.0\%)                             & 0.010                                 \\ \cmidrule(l){2-5} 
                           & Mean                    & 0.3 (0.3\%)             & 50.9 (80.2\%)                             & 0.004                                 \\ \midrule

\multirow{5}{*}{\begin{tabular}[c]{@{}c@{}}~SA-MAP\\ ~(ours)\end{tabular}}      & MT             & \textbf{-0.1 (-0.1\%)}  & 51.0 (77.8\%)                             & \textbf{-0.013}                       \\
                           & US             & 0 (0\%)                 & 45.2 (82.1\%)                             & 0                                     \\
                           & SN             & \textbf{-0.8 (-0.9\%)}  & 49.6 (84.4\%)                             & \textbf{-0.012}                       \\
                           & MM             & \textbf{-0.1 (-0.1\%)}  & \textbf{60.4 (80.2\%)}                    & \textbf{-0.012}                       \\ \cmidrule(l){2-5} 
                           & Mean                     & \textbf{-0.3 (-0.3\%)}  & 51.6 (81.1\%)                             & \textbf{-0.004}                       \\ \bottomrule

\end{tabular}}
\vspace{-0.05in}
\caption{SA-MAP results in source-available situation. Note that the detailed version, likes the form of Table~\ref{tab:source_free}, is in the supplementary. The `$\downarrow$' denotes a smaller number giving a better result, and the `$\uparrow$' means the opposite. The best performances are bolded.}
\label{tab:supervised}
\vspace{-0.25in}
\end{table}

\subsection{Result of MAP in Source-Free Situation}
\label{experiment:SF-MAP}
 
\par
\begin{figure*}[t]
  \centering
  \vspace{-0.10in}
   \includegraphics[width=0.90\textwidth]{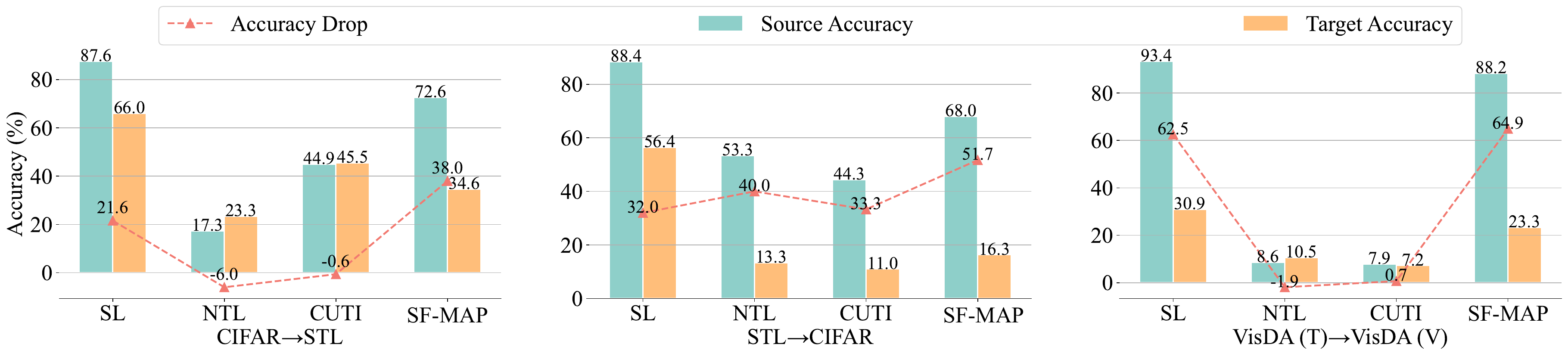}
   \vspace{-0.10in}
   \caption{The accuracy of SL, NTL, CUTI, and SF-MAP on CIFAR10$\rightarrow$STL10, and VisDA-2017 (T$\rightarrow$V). The \textcolor[HTML]{8ECFC9}{green bar}, \textcolor[HTML]{FFBE7A}{orange bar} and \textcolor[HTML]{F27970}{red line} presents the accuracy of the source domain, target domain, and relative degradation, respectively.} 
   \label{fig:cifar_sf}
   \vspace{-0.15in}
\end{figure*}

 \hspace{1em} We then perform model IP protection in the challenging source-free situation and present our SF-MAP solution. We train NTL, CUTI, and SF-MAP with the same samples from Section~\ref{method:sf-ntl} for fairness. Table~\ref{tab:source_free} illustrates a considerable \textit{source drop} of NTL (68.9\%) and CUTI (87.4\%) lead to 1.01 and 1.08 \textit{ST-D}, respectively. While SF-MAP achieves better \textit{Source Drop} (8.8\%) and \textit{ST-D} (0.24). The backbone in Fig.~\ref{fig:cifar_sf} is configured to correspond with the experimental setup described in Section~\ref{experiment:SA-MAP}. In particular, SF-MAP exhibits higher performance with relative degradations of 38.0\%, 51.7\%, and 64.9\%, respectively.

\par We attribute impressive results to the binary mask pruning strategy. Arbitrary scaling of network parameters using continuous masks or direct adjustments has the potential to catastrophic forgetting. Precisely eliminating parameters through a binary mask offers a more effective and elegant solution for IP protection, while concurrently preventing the loss of the network's existing knowledge.

\begin{table*}[]
\vspace{-0.10in}
\centering
\addtolength{\tabcolsep}{4.0pt}
\resizebox{0.98\textwidth}{!}{%
\begin{tabular}{@{}c|c|cccc|cca}
\toprule
\multicolumn{1}{l|}{Methods} & Source/Target & MT          & US          & SN          & MM          & Source Drop$\downarrow$ & \multicolumn{1}{l}{Target Drop$\uparrow$} & \multicolumn{1}{l}{\textit{ST-D}$\downarrow$} \\ \midrule
\multirow{5}{*}{~~~NTL~\cite{wang2021non}}       & MT            & 98.9 / 41.3 & 96.3 / 38.9 & 36.3 / 19.0 & 64.9 / 11.1 & 57.6 (58.2\%)           & 49.5 (65.1\%)                             & 0.89                                  \\
                           & US            & 90.0 / 33.2 & 99.7 / 40.0 & 32.8 / 6.8  & 42.4 / 10.8 & 59.7 (59.9\%)           & 38.1 (69.2\%)                             & 0.86                                  \\
                           & SN            & 68.2 / 20.5 & 75.0 / 32.7 & 92.0 / 19.4 & 32.8 / 9.2  & 72.6 (78.9\%)           & 37.9 (64.5\%)                             & 1.22                                  \\
                           & MM            & 97.5 / 11.3 & 88.3 / 30.7 & 40.2 / 19.0 & 96.8 / 20.6 & 76.2 (78.7\%)           & 55.0 (73.0\%)                             & 1.08                                  \\ \cmidrule(l){2-9} 
                           & Mean          & /           & /           & /           & /           & 66.5 (68.9\%)           & 45.1 (68.0\%)                             & 1.01                                  \\ \midrule
\multirow{5}{*}{~~~CUTI~\cite{wang2023model}}      & MT            & 98.9 / 13.0 & 96.3 / 14.1 & 36.3 / 19.0 & 64.9 / 11.2 & 85.9 (86.9\%)           & \textbf{51.1 (77.5\%)}                    & 1.12                                  \\
                           & US            & 90.0 / 10.7 & 99.7 / 7.8  & 32.8 / 6.6  & 42.4 / 10.6 & 91.9 (92.2\%)           & \textbf{53.1 (83.1\%)}                    & 1.11                                  \\
                           & SN            & 68.2 / 9.3  & 75.0 / 14.1 & 92.0 / 13.6 & 32.8 / 9.4  & 78.4 (85.2\%)           & \textbf{47.7 (81.4\%)}                    & 1.05                                  \\
                           & MM            & 97.5 / 11.4 & 88.3 / 14.1 & 40.2 / 19.0 & 96.8 / 14.2 & 82.6 (85.3\%)           & \textbf{60.5 (80.3\%)}                    & 1.06                                  \\ \cmidrule(l){2-9} 
                           & Mean          & /           & /           & /           & /           & 84.7 (87.4\%)           & \textbf{53.1 (80.6\%)}                    & 1.08                                  \\ \midrule
\multirow{5}{*}{\begin{tabular}[c]{@{}c@{}}~~SF-MAP\\ ~~(ours)\end{tabular}}      & MT            & 99.2 / 90.2 & 96.3 / 59.9 & 36.7 / 19.4 & 64.8 / 24.5 & \textbf{9.0 (9.0\%)}    & 31.3 (47.5\%)                             & \textbf{0.19}                         \\
                           & US            & 90.0 / 67.3 & 99.7 / 83.5 & 32.8 / 7.1  & 42.4 / 30.8 & \textbf{16.2 (16.2\%)}  & 20.0 (36.3\%)                             & \textbf{0.45}                         \\
                           & SN            & 68.2 / 34.8 & 75.0 / 52.5 & 91.4 / 87.2 & 32.8 / 32.0 & \textbf{4.2 (4.6\%)}    & 25.0 (32.2\%)                             & \textbf{0.12}                         \\
                           & MM            & 97.6 / 93.3 & 88.5 / 49.0 & 40.2 / 22.1 & 97.0 / 91.8 & \textbf{5.2 (5.4\%)}    & 19.6 (27.4\%)                             & \textbf{0.20}                         \\ \cmidrule(l){2-9} 
                           & Mean          & /           & /           & /           & /           & \textbf{8.7 (8.8\%)}    & 24.0 (35.9\%)                             & \textbf{0.24}                         \\ \bottomrule

\end{tabular}}
\vspace{-0.1in}
\caption{SF-MAP results in source-free situation. The left of `/' represents the origin source model accuracy with supervised learning, and the right of `/' denotes the accuracy of NTL, CUTI, and SF-MAP trained on the source-free setting. We synthesize samples with source domain features as pseudo-source domains and train with target data. The best performances are bolded.} 
\label{tab:source_free}
\end{table*}

\begin{table*}[]
\vspace{-0.3cm}
\centering
\addtolength{\tabcolsep}{4.0pt}
\resizebox{0.98\textwidth}{!}{%
\begin{tabular}{@{}c|c|cccc|cca}
\toprule
\multicolumn{1}{l|}{Methods} & Source/Target & MT          & US          & SN          & MM          & Source Drop$\downarrow$ & \multicolumn{1}{l}{Target Drop$\uparrow$} & \multicolumn{1}{l}{\textit{ST-D}$\downarrow$} \\ \midrule

\multirow{5}{*}{\begin{tabular}[c]{@{}c@{}}~~DF-MAP\\ ~~(ours)\end{tabular}}      & MT            &  99.1 / 95.0  & 96.8 / 72.0 & 37.1 / 15.1 & 67.5 / 14.7 & 4.1 (4.1\%)   & 33.2 (37.8\%)                          & 0.11                         \\
                           & US            & 89.4 / 83.8 & 99.8 / 99.5 & 34.9 / 31.0 & 33.8 / 16.8 & 0.3 (0.3\%)  & 8.8 (10.3\%)   & 0.03                                                   \\
                           & SN            & 58.6 / 46.6 & 70.9 / 59.2 & 91.5 / 76.3 & 29.5 / 24.2 & 15.2 (16.6\%)    & 9.7 (18.2\%)     & 0.91                         \\
                           & MM            & 98.8 / 94.8 & 84.4 / 61.3 & 40.2 / 28.8 & 96.5 / 94.7 &  1.8 (1.9\%)   & 12.8 (17.2\%)                             &  0.11                        \\ \cmidrule(l){2-9} 
                           & Mean          &  /         & /           & /           & /           & 5.4 (5.7\%)     & 16.1 (20.9\%)                            &  0.27            \\ \bottomrule

\end{tabular}}
\vspace{-0.05in}
\caption{DF-MAP results in the data-free situation. The right of `/' denotes the accuracy of DF-MAP is trained on the data-free setting, which cannot attain any data or labels. The `$\downarrow$' means a smaller number gives a better result, and the `$\uparrow$' means the opposite.} 
\label{tab:data_free}
\vspace{-0.10in}
\end{table*}

\begin{table}[]
\centering
\addtolength{\tabcolsep}{4.0pt}
\resizebox{0.80\linewidth}{!}{
\begin{tabular}{c|cccc}
\toprule
\multirow{2}{*}{Source}                   & \multicolumn{4}{c}{Avg Drop} \\ \cmidrule(l){2-5} 
                              & SL    & NTL   & CUTI  & MAP \\ \midrule
MT                       & 0.2   & 87.9  & 87.7  & 88.6 \\
US                       & 0.1   & 85.7  & 93.0  & 92.5 \\
SN                       & -0.8  & 66.4  & 46.9  & 47.5 \\
MM                       & 4.4   & 83.9  & 79.6  & 79.2 \\
CIFAR                    & 0     & 27.4  & 38.4  & 56.2 \\
STL                      & -6.7  & 54.8  & 62.0  & 60.1 \\
VisDA                    & 0.1   & 0.1   & 22.4  & 19.1 \\ \midrule
Mean                     & -0.3  & 58.0  & 61.4  & 63.3 \\ \bottomrule
\end{tabular}
}
\vspace{-0.10in}
\caption{Ownership verification of SA-MAP. Avg Drop presents the accuracy drop between source domain and watermarked auxiliary domain. Note that the detailed version is in the supplementary.}
\label{tab:ownership}
\vspace{-0.15in}
\end{table}

\subsection{Result of MAP in Data-Free Situation}
  
\par \hspace{1em}We next present our DF-MAP in the extremely challenging data-free setting. Section~\ref{experiment:SF-MAP} indicates that the current techniques are not suited to source-free scenarios. Due to the absence of relevant research to our best knowledge, we refrain from specifying or constructing the baseline in the data-free situation. Table~\ref{tab:data_free} indicates DF-MAP achieved IP protection by achieving a lower decrease in source domains and a higher drop in target domains, as illustrated by \textit{ST-D} being less than 1.0 for all sets of experiments. 

\subsection{Result of Ownership Verification}

\par \hspace{1em} After the above, we additionally conduct an ownership verification experiment of MAP using digit datasets and VGG11 backbone. Following existing work~\cite{wang2021non}, we apply a watermark to source domain samples, treating it as an unauthorized auxiliary domain. As shown in Table~\ref{tab:ownership}, MAP performed 1.9\% better than the second, which demonstrates the utility of this model IP protection approach. 

\subsection{Ablation Study}

\par
\begin{figure}[t]
  \centering
   \includegraphics[width=0.48\textwidth]{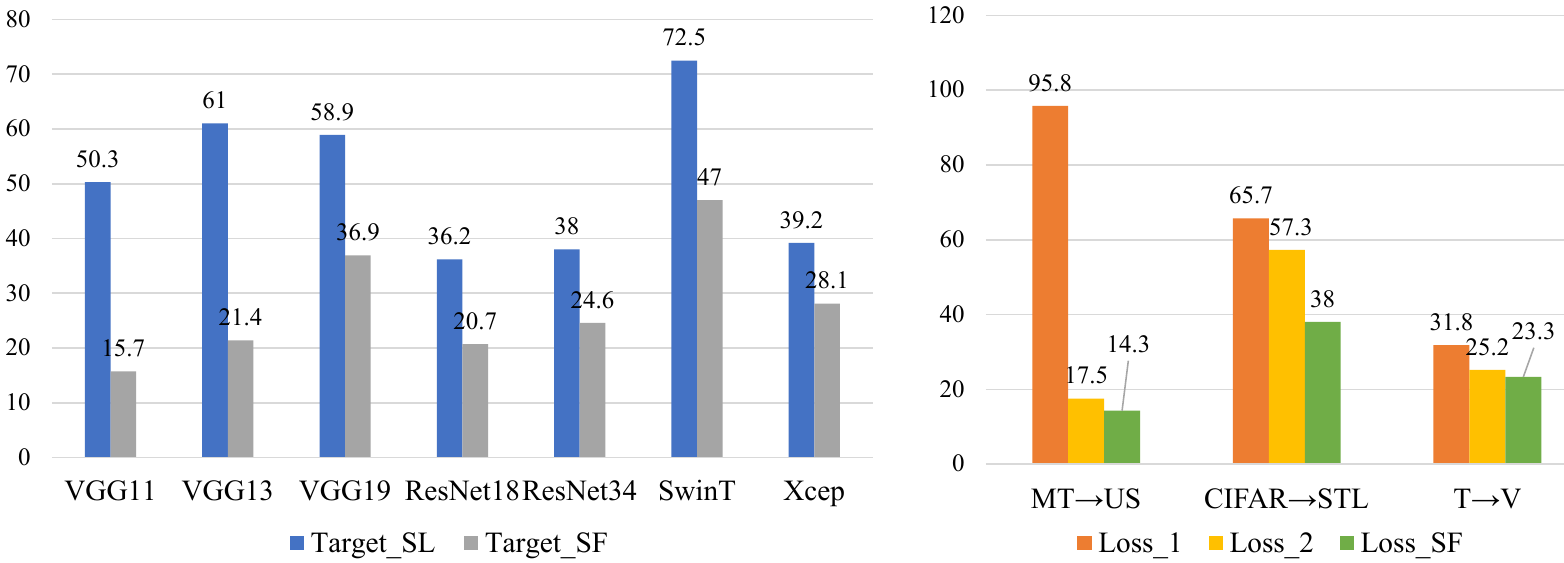}
   \vspace{-0.15in}
   \caption{(a) (\textbf{left}) The accuracy (\%) of origin SL and the SF-MAP model with different backbones on target of STL10 $\rightarrow$ CIFAR10 datasets. (b) (\textbf{right}) The accuracy (\%) of SF-MAP with different losses on the target domain of MT $\rightarrow$ US, CIFAR10 $\rightarrow$ STL10, and VisDA-2017 (T $\rightarrow$ V).} 
   \label{fig:ablation_backbone}
   \vspace{-0.15in}
\end{figure}

 \hspace{1em} \textbf{Backbone.} To verify the generality of MAP for different network architectures, we examine it for several backbones, including VGG11, VGG13, VGG19~\cite{simonyan2014very}, ResNet18, ResNet34~\cite{he2016deep}, Swin-Transformer~\cite{liu2021swin}, and Xception~\cite{chollet2017xception}.
Experiments are conducted on STL10 $\rightarrow$ CIFAR10. We evaluate model accuracy on the target domain with minimal source domain influence. Fig.~\ref{fig:ablation_backbone} (a) illustrates that SF-MAP achieves consistently lower accuracy in unauthorized target domains than the origin model in supervised learning, demonstrating its universality for different backbones.

\par \textbf{Loss Function.} Eq.~(\ref{eq:sf_loss}) suggests $\mathcal{L}_{SF}$ shaped as $\mathcal{L}_1 + \mathcal{L}_2$, where $\mathcal{L}_1 = K L\left(p_s^{\prime\prime} \|~p_t^{\prime\prime}\right)$ and $\mathcal{L}_2 = -K L\left(p_t^T \| y_{psd}\right)$. We conduct ablation studies to verify each loss component's contribution. We utilize $\mathcal{L}_1$, $\mathcal{L}_2$, and $\mathcal{L}_{SF}$ to train SF-MAP on MN$\rightarrow$US, CIFAR10$\rightarrow$STL10, and VisDA-2017 (T$\rightarrow$V). According to Fig.~\ref{fig:ablation_backbone} (b), the result on $\mathcal{L}_1$ has the minimum drop to the target domain due to poor simulation of its features. The result on $\mathcal{L}_{SF}$ shows the greatest target drop without affecting the source domain, but $\mathcal{L}_2$ significantly degrades source performance, as detailed in the supplementary. The $\mathcal{L}_{SF}$ is more accurate than $\mathcal{L}_2$ in recognizing domain-invariant characteristics, eliminating redundant parameters, and degrading the target domain performance.

\par
\begin{figure}[t]
  \centering
  \includegraphics[width=0.49\textwidth]{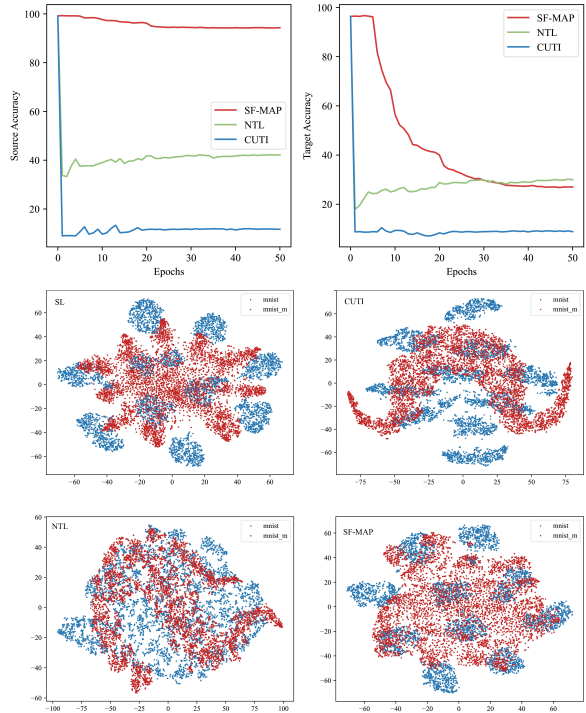}
  \vspace{-0.28in}
  \caption{Visualization analysis. (a) (\textbf{top}) Converge analysis diagram. The convergence of the source accuracy and the target accuracy in the training process is exhibited. (b) (\textbf{bottom}) T-SNE visualization diagram of SL, NTL, CUTI, and SF-MAP.}
  \vspace{-0.20in}
  \label{fig:visualization}
\end{figure}

 \textbf{Visualization.} Fig.~\ref{fig:visualization} (a) illustrates the MN$\rightarrow$US experiment convergence analysis diagram. With SF-MAP, source, and target domain model performance is more balanced. The fact that NTL changes all model parameters may lead to forgetting the source feature and inferior results. In the absence of the real source domain, CUTI's middle domain may aggravate forgetting origin source features since synthesized source domains may have unobserved style features. Fig.~\ref{fig:visualization} (b) illustrates the T-SNE figures of MN$\rightarrow$MM. The origin supervised learning (SL) model, NTL, CUTI, and SF-MAP results are exhibited with the source domain MNIST in blue and the target domain MNISTM in red. As illustrated in Fig.~\ref{fig:visualization} (b), SF-MAP's source domain data retains better clustering information than other approaches, while the target domain is corrupted.

\section{Conclusion}
\label{conclusion}
\par \hspace{1em} Attacks on neural networks have led to a great need for model IP protection. To address the challenge, we present \textbf{MAP}, a mask-pruning-based model IP protection method stemming from our \textit{Inverse Transfer Parameter Hypothesis}, and its expansion forms (SA-MAP, SF-MAP, and DF-MAP) under source-available, source-free, and data-free conditions. SA-MAP updates a learnable binary mask to prune target-related parameters. Based on SA-MAP, SF-MAP uses replay-based generation to synthesize pseudo source samples. We further suggest a diversity generator in DF-MAP to construct neighborhood domains with unique styles. To trade off source and target domains' evaluation, the \textit{ST-D} metric is proposed. Experiments conducted on digit datasets, CIFAR10, STL10, and VisDA, demonstrate that MAP significantly diminishes model generalization region in source-available, source-free, and data-free situations, while still maintaining source domain performance, ensuring the effectiveness of model IP protection. 

\noindent\textbf{Acknowledgment:} This work is supported by the National Natural Science Foundation of China (No.62372329), in part by the National Key Research and Development Program of China (No.2021YFB2501104), in part by Shanghai Rising Star Program (No.21QC1400900), in part by Tongji-Qomolo Autonomous Driving Commercial Vehicle Joint Lab Project, and Xiaomi Young Talents Program.
\clearpage
\setcounter{page}{1}
\maketitlesupplementary

\appendix
\section{Theoretical Analysis}

\par \hspace{1em}Formally, we consider a source network $f_s:\mathcal{X}_s\rightarrow \mathcal{Y}_s$ trained on source domain $\mathcal{D}_s=\{(x_s,y_s)||x_s\sim \mathcal{P}_\mathcal{X}^s, y_s\sim \mathcal{P}_\mathcal{Y}^s\}$, a target network $f_t:\mathcal{X}_t\rightarrow \mathcal{Y}_t$, and target domain $\mathcal{D}_t=\{(x_t,y_t)||x_t\sim \mathcal{P}_\mathcal{X}^t, y_t\sim \mathcal{P}_\mathcal{Y}^t\}$. $\mathcal{P}_\mathcal{X}$ and $\mathcal{P}_\mathcal{Y}$ are the distribution of $\mathcal{X}$ and $\mathcal{Y}$, respectively. The goal of \textbf{\textit{IP protection}} is to fine-tune $f_t$ while minimizing the generalization region of $f_t$ on target domain $\mathcal{D}_t$, in other words, degrade the performance of $f_t$ on unauthorized target domain $\mathcal{D}_t$ while preserving performance on authorized source $\mathcal{D}_s$.

\subsection{Definitions}

\begin{proposition}[\cite{wang2021non}]
Let $n$ be a nuisance for input $x$. Let $z$ be a representation of $x$, and the label is $y$. The Shannon Mutual Information (SMI) is presented as $I(\cdot)$. For the information flow in representation learning, we have
\begin{equation}
I(z ; x)-I(z ; y|n) \geq I(z ; n)
\end{equation}
\label{proposition_1}
\end{proposition}
\vspace{-0.30in}

\begin{lemma}[\cite{wang2021non}]
Let $p$ be the predicted label outputted by a representation model when feeding with input $x$, and suppose that $p$ is a scalar random variable and $x$ is balanced on the ground truth label $y$. And $\mathcal{P}(\cdot)$ is the distribution. If the KL divergence loss $K L(\mathcal{P}(p)\|\mathcal{P}(y))$ increases, the mutual information $I(z; y)$ will decrease.
\label{lemma_1}
\end{lemma}

\subsection{Details of Optimization Objective Design}
\label{sec:objective}

\par \hspace{1em} In the context of intellectual property (IP) protection, the objective is to maximize $I(z; n)$ on the unauthorized domain, and Proposition~\ref{proposition_1} provides guidance by aiming to minimize $I(z; y|n)$. According to Lemma~\ref{lemma_1}, if the Kullback-Leibler (KL) divergence loss $KL(\mathcal{P}(p)\|\mathcal{P}(y))$ increases, the mutual information $I(z ; y)$ will decrease. Since $I(z;y|n)=I(z;y)-I(z;n)$, the $I(z;y|n)$ will consistently decrease with $I(z;y)$. Please note that Proposition~\ref{proposition_1} and Lemma~\ref{lemma_1} have been proved in~\cite{wang2021non}. Therefore, the $\mathcal{L}_O$ in Eq.~\ref{eq:owner_loss}, $\mathcal{L}_{SA}$, and $\mathcal{L}_{SF}$ in the main paper are designed in the form of $\mathcal{L}_1 + \mathcal{L}_2$, where $\mathcal{L}_1 = KL(p_s \|y_s)$ and $\mathcal{L}_2 = -KL(p_t \| y_t)$. This design allows us to maximize $I(z; n)$ on the unauthorized (target) domain and minimize $I(z; n)$ on the authorized (source) domain.

\begin{figure}[t]
  \centering
  
   \includegraphics[width=\columnwidth]{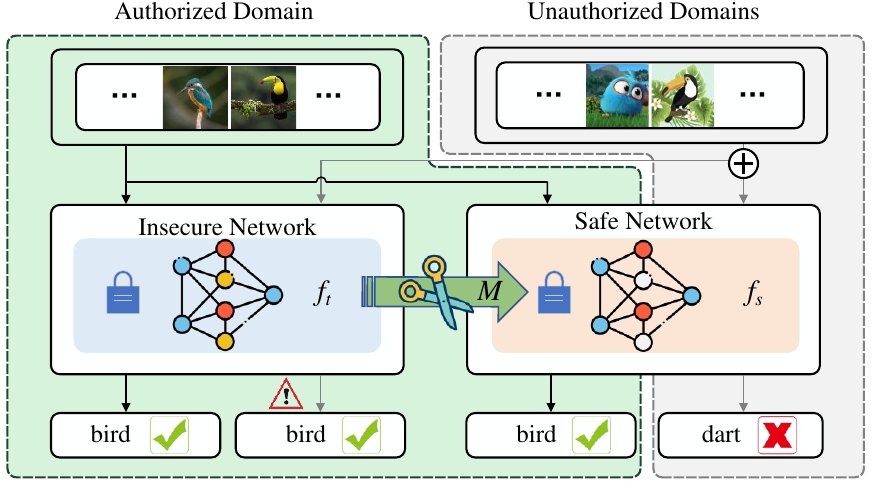}
   \vspace{-0.20in}
   \caption{The total architecture of SA-MAP. A well-trained original source network $f_s$ distills knowledge into the target network $f_t$, which shares the same architecture. We initialize and fix them with the same checkpoint, then update a \textit{Learnable Binary Mask ($M$)} with consistency loss calculated from synthetic samples. The MAP limits a target domain generalization region while retaining source domain performance, leading to a beneficial outcome.}
   \label{fig:sa_map_appendix}
   \vspace{-0.15in}
\end{figure}
\begin{figure*}[t]
  \centering
  \vspace{-0.20in}

   \includegraphics[width=\textwidth]{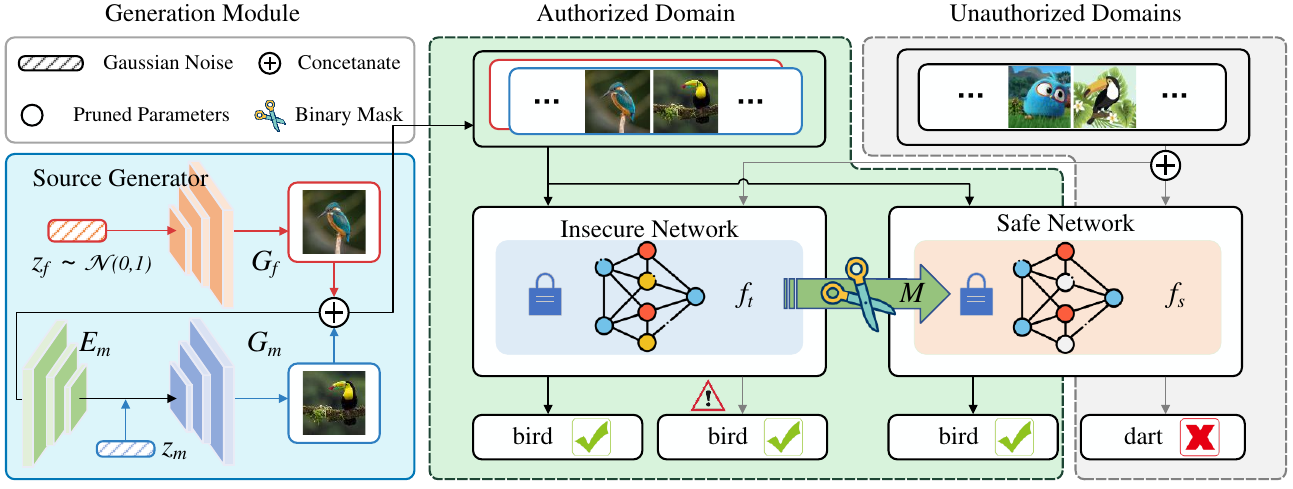}
   \vspace{-0.20in}
   \caption{The total architecture of SF-MAP. \textbf{(a)} The Source Generator, displayed in the left part, consists of two generators. The \textit{Fresh Generator ($G_f$)} generates synthetic novel featured samples, while \textit{Memory Generator ($G_m$)} replays samples with features from previous images. \textbf{(b)} The right part illustrates the mask-pruning process. A well-trained original source network $f_s$ distills knowledge into the target network $f_t$, which shares the same architecture. We initialize and fix them with the same checkpoint, then update a \textit{Learnable Binary Mask ($M$)} with consistency loss from synthetic samples. The MAP limits the target generalization region, leading to a beneficial outcome.}
   \label{fig:sf_map_appendix}
   \vspace{-0.08in}
\end{figure*}

\section{Details of MAP Architecture}
\par \hspace{1em} As elaborated in the main text, we present the comprehensive architectural depiction of DF-MAP. In this supplementary, Fig.~\ref{fig:sa_map_appendix} and Fig.~\ref{fig:sf_map_appendix} showcase the exhaustive architectures of SA-MAP and SF-MAP, respectively. In the source-available scenario, access is available to labeled source samples $\{x_s^i,y_s^i\}_{i=1}^{N_s}$ and target samples $\{x_t^i,y_t^i\}_{i=1}^{N_t}$. We designate the source domain as the authorized domain, anticipating good performance, and the target domain as the unauthorized domain, expecting the opposite. As depicted in Fig.~\ref{fig:sa_map_appendix}, the classification example illustrates that the secure network should correctly classify results on the authorized domain while producing erroneous results on unauthorized ones. We iteratively update a binary mask $M(\theta_M)$ for the insecure model to prune redundant parameters, which contributes to the generalization to unauthorized domains.

\par In the preceding discussion, $f_t$ represents the target network, and $f_s$ represents the source network, both sharing the same architecture and well-trained weight initialization. In the right part of Fig.~\ref{fig:sa_map_appendix}, the pruned source network $f_s$ can be regarded as a sub-network, aligning with our \textit{Inverse Transfer Parameter Hypothesis} outlined in the main paper. More precisely, we gradually update $M(\theta_M)$ to prune redundant target-featured parameters of $f_t$, enabling the pruned $f_t$ to progressively approximate the sub-network of $f_s$. The complete source network $f_s$ is employed in the \textit{Generation Module} in Fig.~\ref{fig:sf_map_appendix} to generate the source-style samples.

\par We further explore the SF-MAP architecture for the source-free model IP protection task, as illustrated in Fig.~\ref{fig:sf_map_appendix}. In the source-free scenario, access is limited to unlabeled target samples $\{x_t^i\}_{i=1}^{N_t}$ and the well-trained weights ($\theta_s$) of $f_s$ trained on the source domain $\mathcal{D}_s$. Consequently, we generate pseudo source domain samples $\{x_s^{i\prime}, y_s^{i\prime}\}_{i=1}^{N_s^{\prime}}$ and pseudo labels $\{y_{psd}^i\}_{i=1}^{N_t}$ for the target domain, as aforementioned. Please note that we are trying to improve the quality of $y_{psd}$, rather than creating a uniform distribution label space for the target domain. Because we want to precisely increase $I(z,n)$ on the target domain, the latter will have a negative influence on the feature $z$ on other domains or even the source domain. Overall, the $y_{psd}$ and \textit{Generation Module} in Fig.~\ref{fig:sf_map_appendix} play a pivotal component within SF-MAP, particularly considering the challenge of synthesizing source-featured samples with only $f_s$ and $\theta_s$.

\section{Details of Experiments}
\subsection{Details of Datasets}

\par \hspace{1em} We implemented our methods and baseline on seven popular benchmarks widely used in domain adaptation and domain generalization. The digit benchmarks include MNIST~\cite{deng2012mnist}, USPS~\cite{hull1994database}, SVHN~\cite{netzer2011reading}, and MNIST-M~\cite{ganin2016domain}. These benchmarks aim to classify each digit into one of ten classes (0-9). MNIST consists of 28x28 pixel grayscale images of handwritten digits, with a training set of 60,000 examples and a test set of 10,000 examples. USPS is a dataset scanned from envelopes, comprising 9,298 16x16 pixel grayscale samples. SVHN contains 600,000 32x32 RGB images of printed digits cropped from pictures of house number plates. MNIST-M is created by combining the MNIST with randomly drawn color photos from the BSDS500~\cite{yang2016object} dataset as a background, consisting of 59,001 training images and 90,001 test images.

Additionally, we employed CIFAR10~\cite{krizhevsky2009learning}, STL10~\cite{coates2011analysis}, and VisDA2017~\cite{peng2017visda} for image classification tasks. CIFAR10 is a subset of the Tiny Images dataset, featuring 60,000 32x32 color images with 10 object classes. STL-10 is an image dataset processed from ImageNet, comprising 13,000 96x96 pixel RGB images with 10 object classes. VisDA2017 is a simulation-to-real dataset for domain adaptation, containing 12 categories and over 280,000 images.

\subsection{Details of SA-MAP result}

\par \hspace{1em} Due to space constraints in the main paper, only a portion of the SA-MAP results are presented. In this supplementary section, we provide the detailed results in Table~\ref{tab:supervised_appendix}. The detailed version primarily includes the drop rate for each experiment group. The results of MAP in the source-available setting exhibit better performance. The true advantage of SA-MAP lies in the elimination of the need for retraining from scratch. The obtained sub-network effectively demonstrates our \textit{Inverse Transfer Parameter Hypothesis}.

\begin{table*}[]
\centering
\vspace{-0.20in}
\addtolength{\tabcolsep}{4.0pt}
\resizebox{0.98\textwidth}{!}{%
\begin{tabular}{@{}c|c|cccc|cca}
\toprule
\multicolumn{1}{l|}{Model} & Source/Target & MT          & US          & SN          & MM          & Source Drop$\downarrow$ & \multicolumn{1}{l}{Target Drop$\uparrow$} & \multicolumn{1}{l}{ST-D$\downarrow$} \\ \midrule
\multirow{5}{*}{NTL}       & MT            & 98.9 / 97.4 & 96.3 / 14.0 & 36.3 / 19.0 & 64.9 / 11.2 & 1.5 (1.5\%)             & 50.9 (77.6\%)                               & 0.019                                  \\
                           & US            & 90.0 / 10.8 & 99.7 / 99.9 & 32.8 / 7.1  & 42.5 / 8.5  & \textbf{-0.2 (-0.2\%)}  & \textbf{46.3 (84.0\%)}                    & \textbf{-0.024}                       \\
                           & SN            & 68.4 / 9.0  & 74.9 / 8.1  & 91.9 / 91.1 & 32.8 / 9.0  & 0.8 (0.9\%)             & \textbf{50.0 (85.2\%)}                    & 0.011                                 \\
                           & MM            & 97.6 / 11.3 & 88.2 / 16.4 & 40.1 / 19.2 & 96.8 / 95.1 & 2.0 (2.1\%)             & 59.7 (79.2\%)                             & 0.027                                 \\ \cmidrule(l){2-9} 
                           & Mean          & /           & /           & /           & /           & 1.0  (1.1\%)            & \textbf{51.7 (81.5\%)}                    & 0.013                                 \\ \midrule
\multirow{5}{*}{CUTI}      & MT            & 98.9 / 98.9 & 96.3 / 7.8  & 36.3 / 19.1 & 64.9 / 12.7 & 0 (0\%)                 & \textbf{52.7 (80.0\%)}                    & 0                                     \\
                           & US            & 90.0 / 16.7 & 99.7 / 99.8 & 32.8 / 10.1 & 42.5 / 8.5  & -0.1 (-0.1\%)           & 42.3 (78.6\%)                             & -0.013                                \\
                           & SN            & 68.4 / 9.3  & 74.9 / 12.6 & 91.9 / 91.6 & 32.8 / 9.2  & 0.3 (0.3\%)             & 48.3 (82.3\%)                             & 0.036                                 \\
                           & MM            & 97.6 / 11.6 & 88.2 / 14.1 & 40.1 / 19.8 & 97.1/96.3   & 0.8 (0.8\%)             & 60.1 (80.0\%)                             & 0.010                                 \\ \cmidrule(l){2-9} 
                           & Mean          & /           & /           & /           & /           & 0.3 (0.3\%)             & 50.9 (80.2\%)                             & 0.004                                 \\ \midrule

\multirow{5}{*}{\begin{tabular}[c]{@{}c@{}}MAP\\ (ours)\end{tabular}}      & MT            & 98.9 / 99.0 & 96.3 / 14.3 & 36.3 / 18.9 & 64.9 / 10.7 & \textbf{-0.1 (-0.1\%)}  & 51.0 (77.8\%)                             & \textbf{-0.013}                       \\
                           & US            & 90.0 / 11.0 & 99.7 / 99.7 & 32.8 / 7.8  & 42.5 / 10.8 & 0 (0\%)                 & 45.2 (82.1\%)                             & 0                                     \\
                           & SN            & 68.4 / 9.5  & 74.9 / 8.5  & 91.9 / 92.7   & 32.8 / 9.4  & \textbf{-0.8 (-0.9\%)}  & 49.6 (84.4\%)                             & \textbf{-0.012}                       \\
                           & MM            & 97.6 / 11.2 & 88.2 / 14.3 & 40.1 / 19.3 & 97.1 / 97.2   & \textbf{-0.1 (-0.1\%)}  & \textbf{60.4 (80.2\%)}                    & \textbf{-0.012}                       \\ \cmidrule(l){2-9} 
                           & Mean          & /           & /           & /           & /           & \textbf{-0.3 (-0.3\%)}  & 51.6 (81.1\%)                             & \textbf{-0.004}                       \\ \bottomrule

\end{tabular}
}
\caption{Results of SA-MAP in source-available situation. In the table, MNIST, USPS, SVHN, and MNIST-M datasets are abbreviated as MN, US, SN, and MM, separately. The left of '/' represents the accuracy of the model trained on the source domain with \textbf{SL}, and the right of '/' means the accuracy of NTL, CUTI, and MAP, which are trained on the \textbf{SL} setting. The 'Source/Target Drop' means the average degradation (relative degradation) of the above models. The '$\downarrow$' means a smaller number gives a better result, and the '$\uparrow$' means the opposite. The data of the NTL and CUTI are obtained by their open-source code. Finally, we bold the number with the best performance.}
\label{tab:supervised_appendix}
\vspace{-0.05in}
\end{table*}

\begin{table*}[t]
\centering
\addtolength{\tabcolsep}{4.0pt}
\resizebox{0.80\textwidth}{!}{%
\begin{tabular}{c|cccc|cccc}
\midrule
\multirow{2}{*}{Source} & \multicolumn{4}{c|}{Methods}                  & \multicolumn{4}{c}{Avg Drop} \\ \cmidrule{2-9} 
                        & SL        & NTL       & CUTI      & MAP      & SL    & NTL   & CUTI  & MAP \\ \midrule
MT                      & 99.2 / 99.4 & 11.2 / 99.1 & 11.4 / 99.1 & 9.7 / 98.3  & 0.2   & 87.9  & 87.7  & 88.6 \\
US                      & 99.5 / 99.6 & 14.0 / 99.7 & 6.8 / 99.8  & 6.8 / 99.3  & 0.1   & 85.7  & 93.0  & 92.5 \\
SN                      & 91.1 / 90.3 & 24.0 / 90.4 & 43.5 / 90.4 & 34.8 / 82.3 & -0.8  & 66.4  & 46.9  & 47.5 \\
MM                      & 92.1 / 96.5 & 12.8 / 96.7 & 17.1 / 96.7 & 16.7 / 95.9 & 4.4   & 83.9  & 79.6  & 79.2 \\
CIFAR                   & 85.7 / 85.7 & 56.8 / 84.2 & 45.3 / 83.7 & 23.5 / 79.7 & 0     & 27.4  & 38.4  & 56.2 \\
STL                     & 93.2 / 86.5 & 26.4 / 81.2 & 22.7 / 84.7 & 22.1 / 82.2 & -6.7  & 54.8  & 62.0  & 60.1 \\
VisDA                   & 92.4 / 92.5 & 93.1 / 93.2 & 73.8 / 92.9 & 70.6 / 89.7 & 0.1   & 0.1   & 22.4  & 19.1 \\ \midrule
Mean                    & /         & /         & /         & /         & -0.3  & 58.0  & 61.4  & 63.3 \\ \midrule
\end{tabular}}
\vspace{-0.05in}
\caption{Ownership verification. The left of '/' denotes NTL, CUTI, and MAP's results on watermarked auxiliary domains, while the right on source domains. The average drop (Avg Drop) presents the drop between source domains and auxiliary domains, the higher, the better.}
\vspace{-0.05in}
\label{tab:ownership_appendix}
\end{table*}

\subsection{Details of Data-Free Model IP Protection}
\par \hspace{1em} We provide the Algorithm~\ref{alg:diversity} of Section~\ref{method:df-ntl} in the main paper. The optimization procedure adheres to the gradient as it represents the most effective path towards achieving the specified objective. In this particular instance, all produced domains exhibit alignment with a consistent gradient direction~\cite{wang2021non}. To introduce diversity in directional perspectives within the generated domains, we impose constraints on the gradient. Specifically, we decompose the generator network $G_d$ into $n_{dir}$ segments. To restrict the direction indexed by $i$, we employ a freezing strategy by fixing the initial $i$ parameters of convolutional layers. This approach entails the immobilization of the gradient with respect to the convolutional layer parameters during training, thereby constraining the model's learning capacity along that particular direction.
\begin{algorithm}[t]
\caption{Diversity Neighborhood Domains Generation}
\label{alg:diversity}
\begin{algorithmic}[1]
\REQUIRE The input samples $\mathcal{X}$ and label $\mathcal{Y}$ diversity generator network $G_d(x;\theta_{\mu}, \theta_{\sigma})$, direction number $n_{dir}$, neighborhood samples $\mathcal{X}_{nbh}=[~]$
\WHILE{not converged}
    \FOR{$d$ in $n_{dir}$}
        \STATE Generate sample $x_g$ = $G_d(x)$
        \STATE Freeze the first $d$ parts of $G_d$'s each layers 
        \STATE Build MI loss $\mathcal{L}_{MI}$ of $x_g$ and $x$ as Eq.~(\ref{eq:mutual_information})
        \STATE Build semantic loss $\mathcal{L}_{sem}$ of $x_g$, $x$ as Eq.~(\ref{eq:semantic_consistency})
        \STATE Update $\theta_{\mu}$ and $\theta_{\sigma}$ by $\mathcal{L}_{MI}+\mathcal{L}_{sem}$
    \ENDFOR
    \STATE Append $x_g$ to $\mathcal{X}_{nbh}$  
\ENDWHILE
\RETURN Neighborhood samples $\mathcal{X}_{nbh}$
\end{algorithmic}
\end{algorithm}

\subsection{Details of Ownership Verification}

\par \hspace{1em} We adopt the method from~\cite{wang2021non} to introduce a model watermark to the source domain data, creating a new auxiliary domain $\mathcal{D}_a=\{(x_a,y_a)||x_a\sim \mathcal{P}_\mathcal{X}^a, y_a\sim \mathcal{P}_\mathcal{Y}^a\}$. In this experiment, we utilize the watermarked auxiliary samples $\{x_a^i,y_a^i\}_{i=1}^{N_a}$ as the unauthorized samples, exhibiting poor performance when evaluated by model $f_t$. The original source samples $\{x_s^i,y_s^i\}_{i=1}^{N_s}$ without watermarks represent the authorized samples, intended to yield good performance. The details of the ownership verification experiments are outlined in Algorithm~\ref{alg:ownership}.

\begin{figure*}[t]
  \centering
   \includegraphics[width=0.80\textwidth]{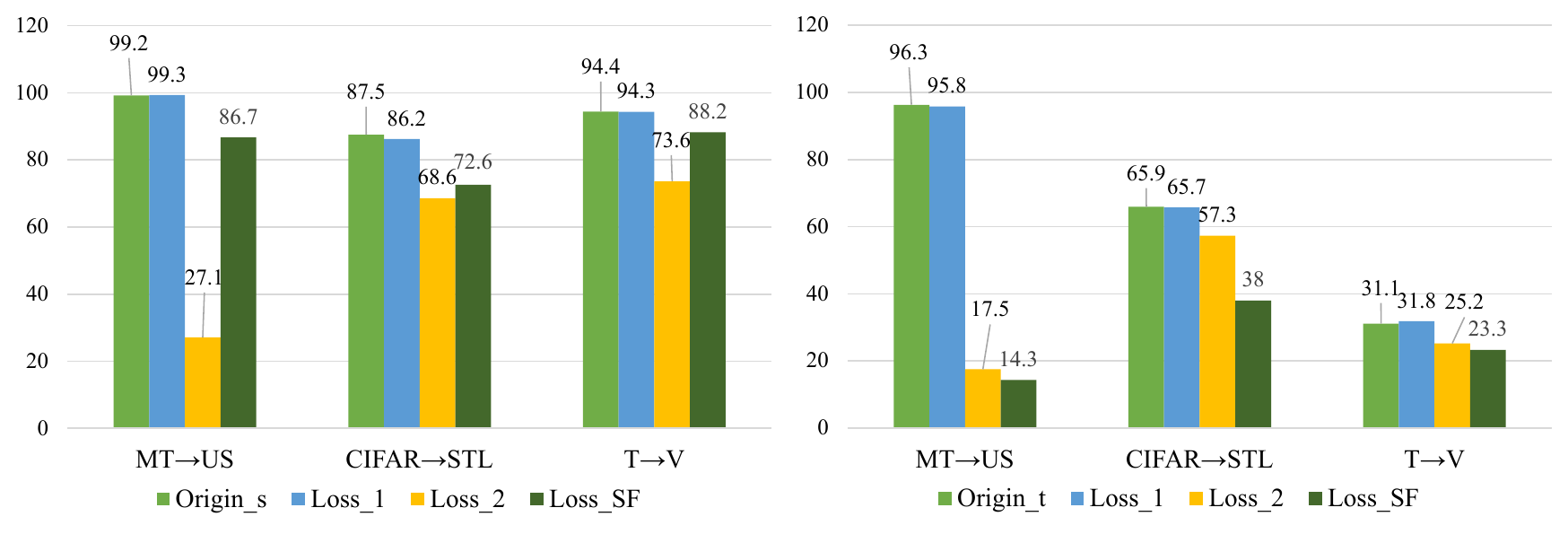}
   \vspace{-0.010in}
   \caption{The accuracy (\%) of SF-MAP with different losses on the target domain of MT $\rightarrow$ US, CIFAR10 $\rightarrow$ STL10, and VisDA-2017 (T $\rightarrow$ V). The \textbf{left} sub-figure is the performance of the source domain, and the \textbf{right} is the performance of the target domain. The \textcolor[HTML]{70AD47}{light green bar}, \textcolor[HTML]{5B9BD5}{blue bar}, \textcolor[HTML]{FFC000}{yellow bar}, and the \textcolor[HTML]{43682B}{dark green bar} present the accuracy of origin model, the accuracy of $\mathcal{L}_1$-trained model, the accuracy of $\mathcal{L}_2$-trained model, and the accuracy of $\mathcal{L}_{SF}$-trained model, respectively.}
   \label{fig:loss_appendix}
\end{figure*}

\vspace{-0.25in}
\begin{equation}
\begin{aligned}
\mathcal{L}_{O}(f_t;\mathcal{X}_s,\mathcal{Y}_s, \mathcal{X}_a, \mathcal{Y}_a)= \frac{1}{N_s} \sum_{i=1}^{N_s} KL(p_t^S \| y_s) \\- min\{\lambda \cdot\frac{1}{N_a} \sum_{i=1}^{N_a} KL(p_t^A \| y_a), \gamma\}
\end{aligned}
\label{eq:owner_loss}
\end{equation}
where $K L(\cdot)$ presents the Kullback-Leibler divergence. $p_t^S = f_t(x_s)$ and $p_t^A = f_t(x_a)$ mean the prediction of target model $f_t$. $\lambda$ means a scaling factor and $\gamma$ means an upper bound. We set $\lambda = 0.1$ and $\gamma = 1$

\par The detailed results in Table~\ref{tab:ownership_appendix} reveal that the original model in supervised learning (SL) struggles to differentiate between the source domain $\mathcal{D}_s$ and the watermarked authorized domain $\mathcal{D}_a$, achieving similar results on both. In contrast, established model intellectual property (IP) protection methods such as NTL~\cite{wang2021non}, CUTI~\cite{wang2023model}, and our SA-MAP exhibit distinct advantages. These methods showcase superior performance on $\mathcal{D}_s$ compared to $\mathcal{D}_a$. Particularly noteworthy is the performance of SA-MAP, which outperforms the other methods, showcasing a 1.9\% improvement over the second-best approach.
\begin{algorithm}[t]
\caption{Ownership Verification with MAP}
\label{alg:ownership}
\begin{algorithmic}[1]
\REQUIRE The source dataset $\mathcal{X}_s$, target model $f_t(x;\theta_t)$, pre-trained model parameters $\theta_0$, mask $M(\theta_M)$. 
\STATE Initialize $\theta_t$ with $\theta_0$ and fix $\theta_t$ 
\WHILE{not converged}
    \STATE Add watermark to $x_s$ to build $x_a$ as~\cite{wang2021non}.
    \STATE Update $\theta_M$ by $x_s$ and $x_a$ as Eq.~(\ref{eq:owner_loss})
\ENDWHILE
\RETURN Learned mask parameters $\theta_M$
\end{algorithmic}
\end{algorithm}

\subsection{Details of Ablation Study}

\par \hspace{1em} The detailed ablation figure, highlighting various loss functions, is depicted in Fig.~\ref{fig:loss_appendix}. As stated before, the model trained with $\mathcal{L}_{SF}$ demonstrates superior performance on both the source and target domains. In the case of $\mathcal{L}_{1}$, the outcome closely resembles that of the original model, indicating a lack of effective model intellectual property (IP) protection. On the other hand, with $\mathcal{L}_{2}$, although there is a reduction in accuracy on the target domain, there is a noteworthy decline on the source domain, signifying a failure to adequately preserve the source performance.
{
    \small
    \bibliographystyle{ieeenat_fullname}
    \bibliography{main}
}


\end{document}